\documentclass[a4paper,11pt]{article}
\usepackage{adjustbox}
\usepackage{booktabs}
\usepackage{xcolor}
\usepackage{multirow}
\usepackage{fourier}
\usepackage{color}
 \usepackage{graphicx}
 \usepackage{csvsimple} 
\usepackage{url}
\usepackage{pdfpages}
\usepackage{subcaption}
\usepackage[affil-it]{authblk}
\usepackage{amsmath}
\usepackage{wrapfig}
\usepackage{times}
\begin{document}
\title{Deformable Convolution Based Road Scene Semantic Segmentation of  Fisheye Images in Autonomous Driving}

\author{\small \textit{\textbf{Anam Manzoor}\textsuperscript{1},
        \textbf{Aryan Singh}\textsuperscript{1},
        \textbf{Ganesh Sistu}\textsuperscript{1},
        \textbf{Reenu Mohandas}\textsuperscript{1},
        \textbf{Eoin Grua}\textsuperscript{1},
        \textbf{Anthony Scanlan}\textsuperscript{1},
        \textbf{Ciarán Eising}\textsuperscript{1}}}

\affil{\small \textit{\textsuperscript{1}Data-Driven Computer Engineering Group, Dept. of Electronic and Computer Engineering, University of Limerick}}

\affil{\bf This paper is a preprint of a paper submitted to the 26th Irish Machine Vision and Image Processing
Conference (IMVIP 2024). If accepted, the copy of record will be available at IET Digital Library.}

\date{}
\maketitle
\thispagestyle{empty}

\vspace{-10mm}
\begin{abstract}
This study investigates the effectiveness of modern Deformable Convolutional Neural Networks (DCNNs) for semantic segmentation tasks, particularly in autonomous driving scenarios with fisheye images. These images, providing a wide field of view, pose unique challenges for extracting spatial and geometric information due to dynamic changes in object attributes. Our experiments focus on segmenting the WoodScape fisheye image dataset into ten distinct classes, assessing the Deformable Networks' ability to capture intricate spatial relationships and improve segmentation accuracy. Additionally, we explore different loss functions to address class imbalance issues and compare the performance of conventional CNN architectures with Deformable Convolution-based CNNs, including Vanilla U-Net and Residual U-Net architectures. The significant improvement in mIoU score resulting from integrating Deformable CNNs demonstrates their effectiveness in handling the geometric distortions present in fisheye imagery, exceeding the performance of traditional CNN architectures. This underscores the significant role of Deformable convolution in enhancing semantic segmentation performance for fisheye imagery.
\end{abstract}
\section{Introduction}
\begin{wrapfigure}{r}{0.5\textwidth}
  \vspace{-20pt}
  \begin{center}
    \includegraphics[width=0.48\textwidth, height=0.28\textwidth]{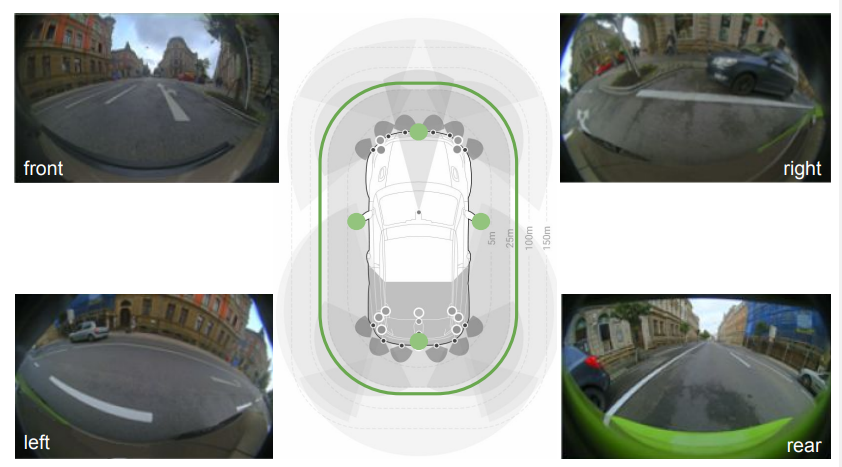}
\end{center}
\vspace{-15pt}
 \captionsetup{font=small}
  \caption{Four fisheye cameras mounted around the vehicle to provide complete 360-degree coverage.}
  \label{fig: Figure 1}
  \vspace{-0pt}
\end{wrapfigure}
Semantic segmentation is indispensable in autonomous driving, as it enables precise object recognition and scene comprehension, essential for safe navigation \cite{divakarla2023semantic}. By accurately labelling pixels in images, vehicles can perceive pedestrians, road lanes, and other critical elements in their surroundings, facilitating informed decision-making and boosting traffic safety.
In autonomous driving, images acquired from multiple fisheye cameras positioned around the vehicle provide a comprehensive 360° view of the surroundings \cite{ramachandran2021woodscape} as shown in Figure \ref{fig: Figure 1}. Unlike conventional cameras, fisheye cameras provide broader scene coverage, which is particularly beneficial in navigating complex urban environments. Thus, integrating semantic segmentation with fisheye images enhances autonomous vehicles' capabilities, allowing them to interpret scenes and navigate adeptly across diverse driving scenarios accurately, as discussed in \cite{cho2023surround}.
However, fisheye images often suffer from the loss of translation equivariance, necessitating specialized CNN architectures capable of effectively handling these translation transformations. Traditional CNN algorithms such as spherical CNNs \cite{cohen2018spherical} and mapped CNNs \cite{eder2019mapped} encounter challenges when applied to fisheye images due to their inherent radial distortion, requiring rectification before application. These rectification methods involve cubic and spherical mapping as illustrated in \cite{wang2018cubemapslam}, \cite{cho2023surround}, and \cite{hawary2020sphere}, respectively, aim to mitigate the effects of fisheye distortion. Yet, this rectification approach presents drawbacks such as resampling distortion artefacts and reduced field-of-view, highlighting the need for innovative solutions. Therefore, after exploring various CNN networks and hypothesizing potential solutions, recent advancements in Deformable Convolution Neural Networks have emerged as promising candidates. As introduced by \cite{Dai_2017_ICCV}, \cite{deng2019restricted}, and \cite{Zhu_2019_CVPR}, Deformable CNNs offer a compelling approach where the shape of the convolutional kernel is dynamically adjusted based on the object's shape and position during training. This adaptability aligns well with the challenges posed by fisheye imagery, potentially enabling more efficient and accurate modelling of spatial relationships within distorted images.

Thus, our study primarily focuses on exploring the effectiveness of Deformable Convolutions as an alternative to regular convolutional layers in segmenting fisheye images using  U-Net \cite{ronneberger2015u}. We examine multi-view scene processing to assess the versatility of our approach. Additionally, we explore how incorporating images from various scenes can generalize the model to improve the segmentation accuracy while evaluating the effectiveness of different versions of the U-Net model. Rather than solely aiming for state-of-the-art results, we highlight the role of Deformable Convolutions in facilitating view-agnostic learning as discussed in \cite{shang2022learning}, also shedding light on their intrinsic advantages for fisheye image segmentation. We also provide baseline results from the  Woodscape dataset to validate our findings.

This paper starts by providing a thorough background study in Section 2, followed by an explanation of proposed methodologies for semantic segmentation of fisheye images in Section 3. Section 4 examines the outcomes from the implemented models, while Section 5 summarizes the paper's conclusions and suggests future research directions.
\section{Background Study on Semantic Segmentation in Automotive Imaging}
A pioneering approach for image segmentation was using CNN networks. Semantic segmentation methods utilizing Convolutional Neural Networks distinguish themselves from other classical techniques owing to their capacity for end-to-end training and robust generalization to novel and limited data. CNNs also play a pivotal role by encapsulating prior knowledge regarding geometric transformations, facilitated by their adaptable model capacity and translational invariant properties, achieved through max-pooling layers within the network \cite{long2015fully}. With the widespread accessibility of high-performance Graphics Processing Units (GPUs) and intuitive deep learning frameworks, CNN-based models have achieved notable advancements across diverse domains, notably within semantic segmentation for autonomous driving \cite{huang2020survey}, and \cite{xie2021segformer}.

The U-Net architecture, extensively utilized for tasks like semantic segmentation and image translation in computer vision, follows an encoder/decoder design with batch-normalization and ReLU activation functions after each convolution block, as detailed in \cite{ronneberger2015u}. Additionally, the Residual U-Net, an evolved variant discussed in \cite{quan2021fusionnet}, incorporates residual blocks and internal long skip connections, along with extra operations between convolution blocks in the encoder and decoder paths, enhancing information flow and mitigating gradient vanishing issues. These architectures, known for their encoder-decoder design, efficiently extract features and generate segmentation maps. Skip connections help preserve spatial information for precise object localization, while the Residual U-Net's integration of residual connections ensures robust network training. Overall, these frameworks excel in semantic segmentation tasks due to their efficient operations, adaptability, and capacity to capture both local and global contexts.

Fisheye cameras, commonly employed in autonomous driving, offer an expanded field of view compared to rectilinear images, capturing a broad spectrum of visual information. This inherent advantage makes fisheye cameras particularly valuable across diverse applications such as intelligent surveillance, drone technology \cite{yang2020intelligent}, and autonomous driving \cite{kumar2021omnidet}. Nevertheless, its unique spherical view introduces significant distortions, where objects near the fisheye lens within a fisheye image exhibit considerable enlargement and distortion. At the same time, those farther away experience larger distortions due to low pixel density on the edges. As a result, the same object appears with different shapes at various positions within fisheye images, posing a unique challenge for the semantic segmentation task \cite{zhou2024improved}.

The semantic segmentation of fisheye images using convolutional neural networks (CNNs) has presented unique challenges compared to rectilinear images. CNNs struggle with handling large-scale spatial transformations due to their fixed receptive fields within the network \cite{Dai_2017_ICCV}. Thus, the optical distortions inherent in fisheye imagery impede CNNs' ability to model unknown geometric transformations effectively. These distortions vary based on the object's view angle relative to the camera, adding complexity to segmentation tasks. While various data augmentation techniques and synthetic data generation methods using regular CNNs have been explored, they often inadequately represent the complexities of real-world fisheye images \cite{saez2018cnn, ye2020universal, deng2017cnn}. Therefore, training a model with raw fisheye data and applying semantic algorithms without undistortion could offer an optimal solution for building a generalized model. Furthermore, our literature review revealed a scarcity of publicly available fully annotated fisheye datasets for understanding road scenes. Only a few datasets, including OmniScape, \cite{sekkat2020omniscape}, WoodScape \cite{yogamani2019woodscape}, and SynWoodScape \cite{sekkat2022synwoodscape}, provide semantic segmentation ground truths specifically designed for fisheye images.

The introduction of Deformable Convolution, followed by the advancements in modulated Deformable convolution introduced by  \cite{Dai_2017_ICCV} and \cite{Zhu_2019_CVPR}, respectively, has significantly enhanced the ability to handle geometric transformations in computer vision tasks. Modulated Deformable Convolution dynamically adapts to the dimensions and contours of detected objects, thereby enabling precise focus on relevant image regions. In contrast to conventional techniques using a fixed position kernel sampling, the deformable convolution introduces learned positional offsets to every sampled point of the kernel. This convolutional mechanism dynamically learns the perceptual field based on identified objects, significantly enhancing spatial sampling and producing robust and accurate feature representation. However, there has been limited exploration into harnessing Deformable Convolutions to address the challenge of learning unfamiliar geometric transformations in real-world fisheye imagery, especially within the automotive sector.

A recent study by \cite{cho2023surround} introduced a novel viewpoint augmentation technique, leveraging the Woodscape dataset to capture the distortion characteristics inherent in fisheye images. Furthermore, to our knowledge, only one recent study has compared regular CNNs with Deformable Convolutions for semantic segmentation of automotive fisheye images by deploying Residual U-Net \cite{el2023fully}, demonstrating promising results in adapting to the unique characteristics of fisheye images with a single view, i.e., only Front view training and testing. To the best of our knowledge, the current literature has not explored the integration of the Deformable Convolutional component into the vanilla U-Net model for fisheye image segmentation, particularly in accommodating diverse viewpoints like Front View, Rear View, and Mirror Left and Right Views during both training and testing phases to enhance model generalization and performance.
\section{Methodology}
Our primary model builds upon the U-Net and Residual U-Net architectures proposed in \cite{ronneberger2015u} and \cite{quan2021fusionnet} respectively, incorporating Deformable Convolutions as discussed in \cite{deng2019restricted}. We expand the model derived from the vanilla U-Net and Fully Residual U-Net architectures by substituting the traditional convolutional blocks with Deformable Convolutions. We focus on presenting only one variation with Deformable Convolution Blocks to limit the total number of experiments. Our primary contribution lies in modifying the initial and final convolutional layers of both the vanilla U-Net and Residual U-Net, as illustrated in Figure \ref{fig: Figure 2}. This alteration enables the network to account for spatial and geometric characteristics during the training process\cite{deng2019restricted}. Moreover, we explore different variants of the baseline U-Net network, namely V\_U-Net, V\_DeU\-Net, R\_U-Net, and R\_DeU-Net  models.
\subsection{Dataset}
To evaluate our approach and establish a correlation between our model's performance and the complexities inherent in fisheye imagery, we utilized the publicly available real-world WoodScape dataset introduced in\cite{yogamani2019woodscape}. This dataset consists of 10,000 annotated images captured from four different view angles: Front View (FV), Mirror-View Right (MVR), Mirror-View Left (MVL), and Rear View (RV) of a vehicle. The dataset provides the semantic annotations for ten classes, including road, lane markings, curb, person, rider, car, bicycle, motorcycle, traffic sign, and background. Upon examining the dataset characteristics, it becomes apparent that a significant class imbalance exists concerning the occurrence of specific classes, such as bicycle, motorcycle, and traffic signs, which are less dominant than the road, landmarks, and curb courses across the entire image dataset.
\vspace{-5mm}
\begin{figure*}
    \centering
    \includegraphics[width=0.8\linewidth]{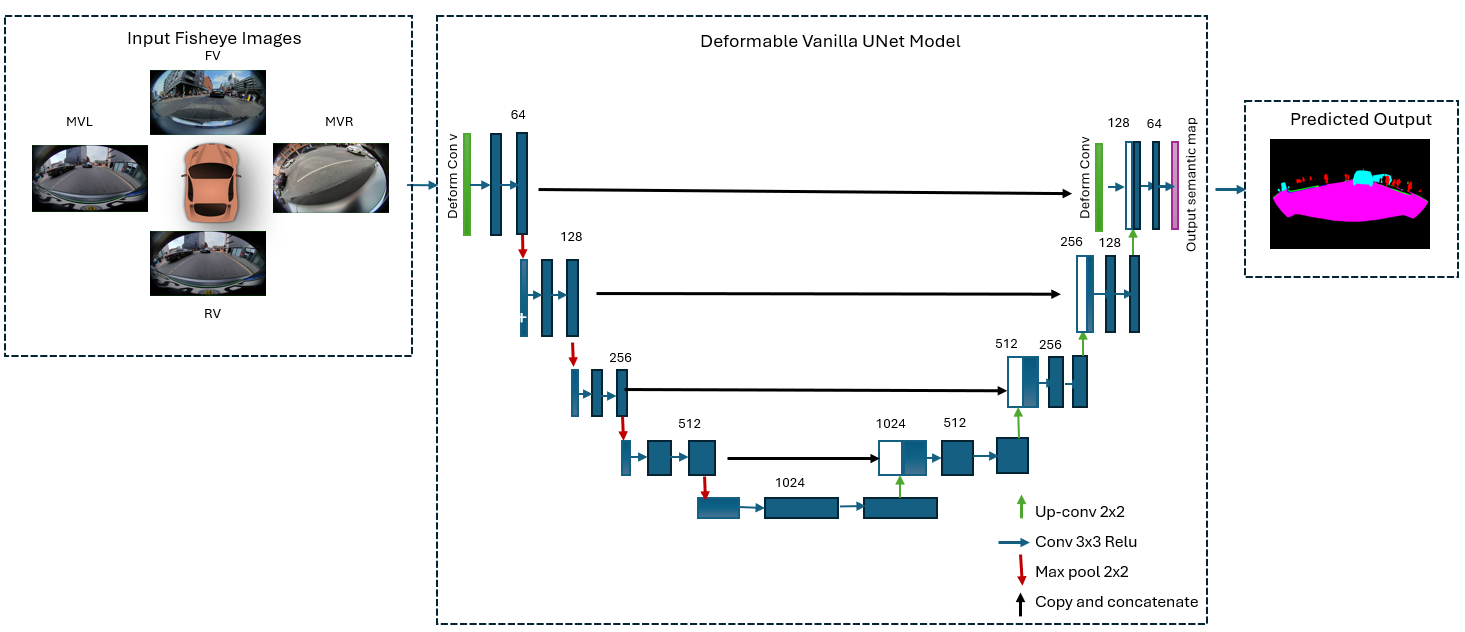}
    \captionsetup{font=small}
    \caption{Baseline Vanilla DeU-Net model where Deformable Convolution block injected into the first layer of the encoder and last layer of decoder path to better account the spatial and geometric characteristics of fisheye images during training.}
    \vspace{-0.5cm}
    \label{fig: Figure 2}
\end{figure*}

\subsection{Experimental Setup}

To ensure reproducibility, we established an experimental framework. We implemented both the vanilla and Fully Residual U-Net model with and without Deformable convolution blocks referred to as V\_U-Net (Vanilla U-Net), V\_DeU\-Net (Vanilla Deformable U-Net), R\_U-Net (Residual U-Net), and R\_DeU-Net (Residual Deformable U-Net) models. Moreover, we explore the effectiveness of the Deformable U-Net model across the multi\_view training and testing instead of a single view. 
\begin{figure}
   \setcounter{figure}{0}
    \centering
    \vspace{-0.1cm}
    \begin{minipage}[b]{0.45\linewidth} 
        \centering
        \includegraphics[width=8cm,height=6cm]{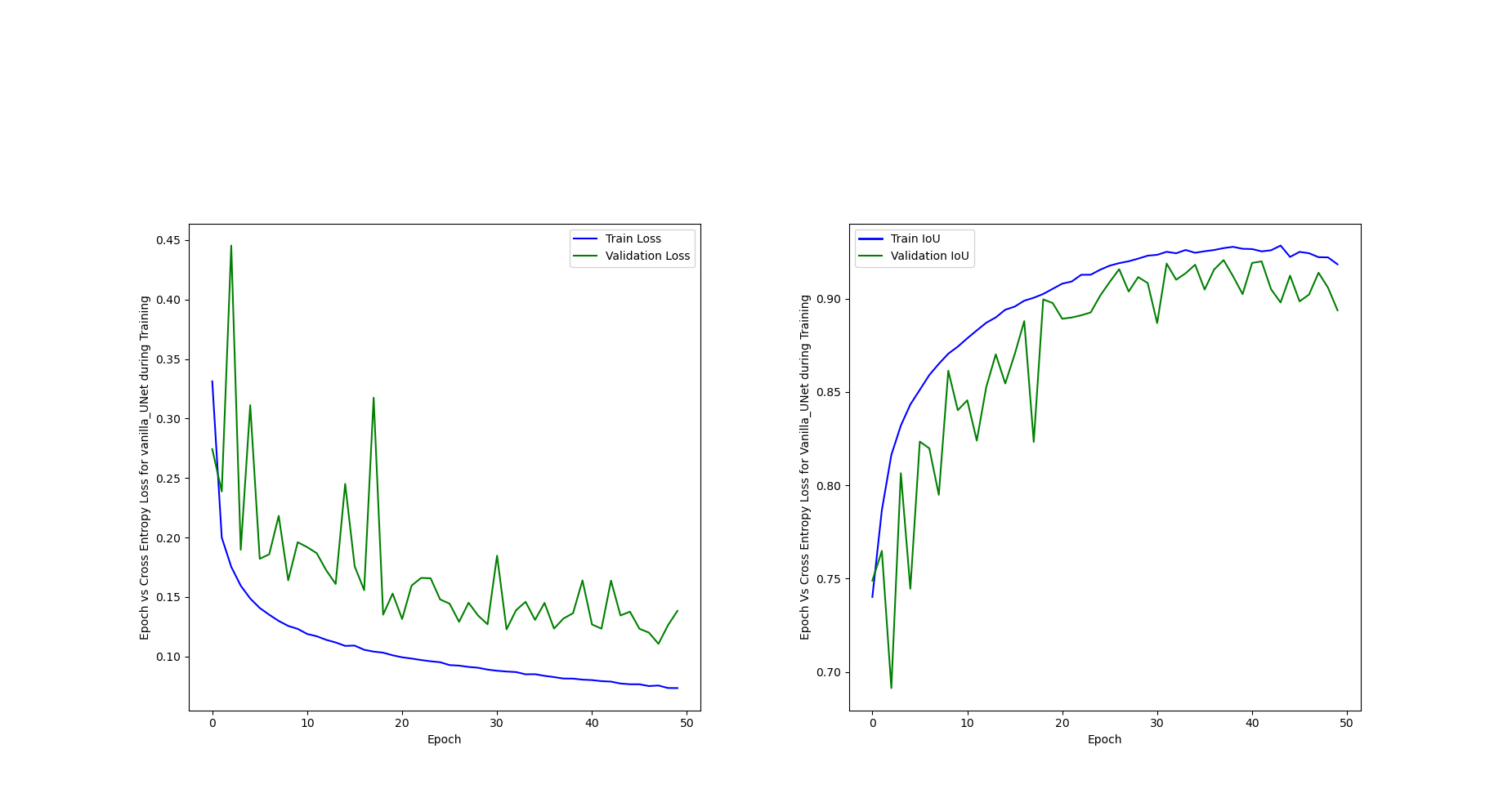}
        \captionsetup{font=small}
        \captionsetup{labelformat=empty}
        \vspace{-0.9cm}
        \caption{(a)}
        \label{fig:sub1}
    \end{minipage}
    \hfill
    \begin{minipage}[b]{0.45\linewidth} 
        \centering
        \includegraphics[width=8cm,height=5cm]{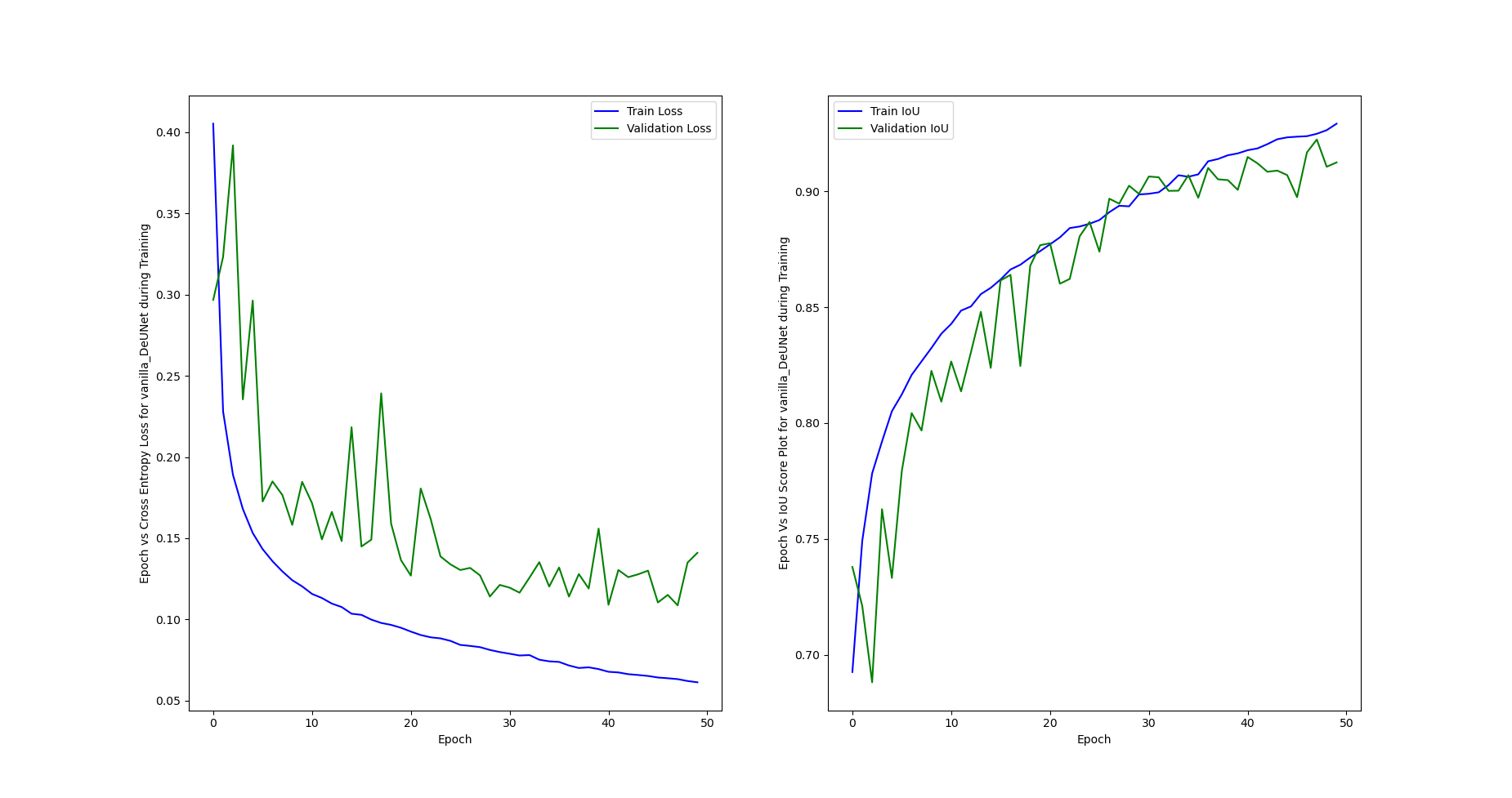}
        \captionsetup{font=small}
        \captionsetup{labelformat=empty}
        \vspace{-0.9cm}
        \caption{(b)}
        \label{fig:sub2}
    \end{minipage}
    \captionsetup{font=small}
    \vspace{-0.3cm}
    \caption{Epoch Vs Cross Entropy Loss and IoU score illustrated by Figures (a) and (b) showing the Epoch Vs Cross Entropy Loss and IoU score with Vanilla U-Net and Vanilla DeU-Net throughout the training process.}
    \label{fig: Figure 3}
\end{figure}
We initially partitioned the datasets into training, validation, and test sets using an 80\% , 10\% , and 10\%  split, respectively. This split ensured that all views were uniformly distributed across the dataset partitions. The training was done from scratch using the PyTorch framework on an NVIDIA GeForce RTX 3080 GPU. We employed the Adam optimizer and initialized the weights randomly, selecting a batch size of 1 to facilitate rapid adaptation of the model to the dataset while mitigating the risk of overfitting to the limited training data and preventing memory errors. The training continued for 50 epochs without encountering any memory issues. The initial learning rate is set to \(1 \times 10^{-4}\) and adjusted according to the validation performance during training.

\begin{figure}
    \centering
    \includegraphics[width=1\linewidth]{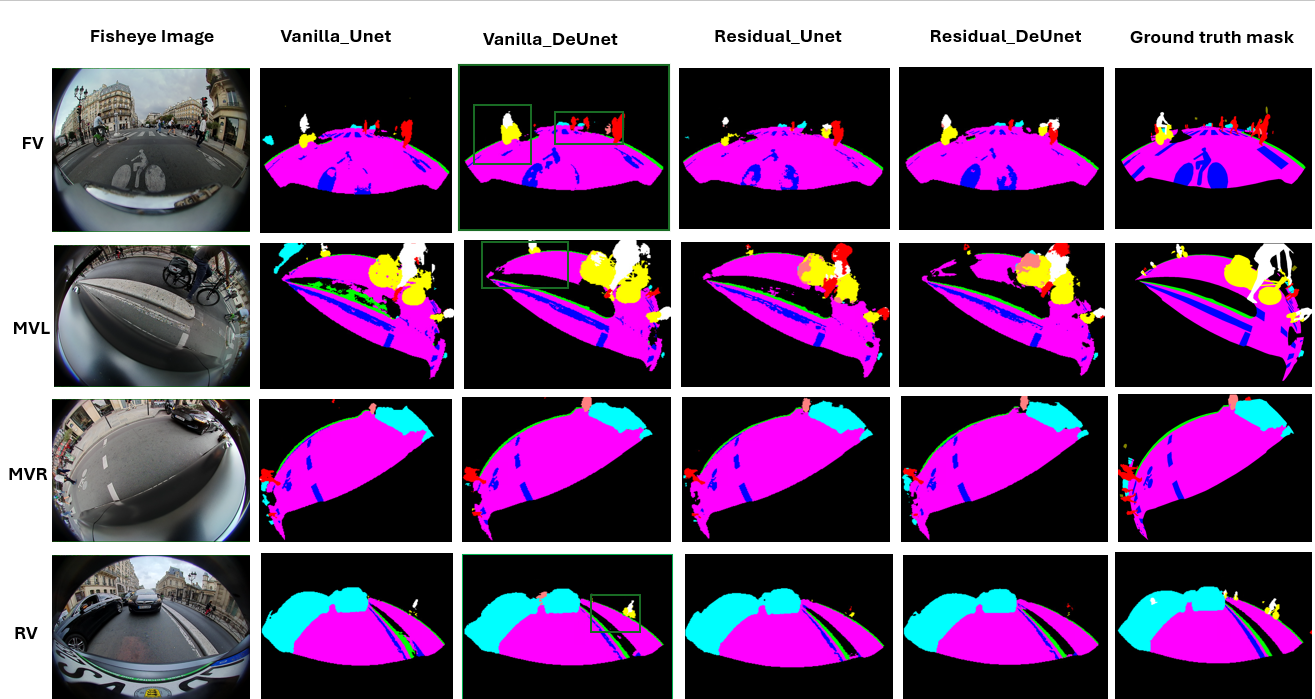}
    \captionsetup{font=small}
    \caption{Visualizations of results on Woodscape fisheye images and corresponding ground truth masks are presented across baseline models including Vanilla\_U-Net, Residual\_U-Net, Deformable\_U-Net, and Deformable\_Residual model. Notably, the visualization performance with Vanilla\_DeU-Net surpasses the other models, as indicated with green boxes compared to ground truth masks, with particular emphasis on distorted edges.}
    \label{fig: Figure 4}
\end{figure}

Moreover, to identify the most effective approach for handling the imbalanced dataset, we experimented with different variants of loss functions, including standard cross-entropy loss, focal loss, and class-weighted focal loss, calculated through a re-weighting scheme as discussed in \cite{paszke2016enet}. 
For pre-processing, we resized the RGB images to dimensions of 256x256 for computation efficacy. During training and validation, we applied augmentation techniques such as horizontal flipping (with a probability of 0.5) with random adjustments to brightness and contrast to improve the model's generalization ability.Additionally, we normalized the pixel values of the images to ensure they fell within the range of 0 to 1 and transformed into tensors. The curves illustrating the Training and Validation Epoch versus Cross-Entropy Loss, as well as IoU scores, are depicted in Figure \ref{fig: Figure 3}. All training curves are excluded due to space constraints. Finally, We evaluated the performance of the models on the test set annotations, which included all four side surround view images and masks, using both the mean Intersection over Union (mIoU) and accuracy metrics. 
The visualization results of four baseline models, Vanilla\_U-Net, Vanilla\_DeU-Net, Residual\_U-Net, and Residual\_DeU-Net, to assess the efficacy of Deformable convolutions on baseline models are shown in Figure \ref{fig: Figure 4}.

\section{Results and Experiments}
Integrating Deformable blocks into the initial layers of the encoder and the final layer of the decoder paths of U-Net variants substantially improved class IoU scores and accuracy compared to the Vanilla and Residual U-Net baseline models. As discussed earlier, we utilized three different loss functions, cross-entropy loss, focal loss, and weighted focal loss, to explore their impact on model performance and the imbalanced dataset. The results, as summarized in Table 1, underscore the significance of our approach for fisheye datasets.
To elaborate further, Table\ref{tab: Table 1} initially presents the outcomes with the baseline Vanilla\_U-Net\_ce, outlining the IOU score for each class. Subsequently, upon integrating Deformable blocks into the baseline Vanilla\_U-Net\_ce (here, "ce" represents training with cross-entropy loss), notable enhancements in IOU scores for specific classes were observed: Curb by 0.01, Person by 0.11, Rider by 0.05, Bicycle by 0.001, Motorcycle by 0.01, and Traffic Sign by 0.10, respectively.

Similarly, when Deformable blocks were integrated into the Residual U-Net network, the IOU scores for each class outperformed those of the baseline Residual\_U-Net. Furthermore, we explored the effectiveness of different loss functions, including standard and weighted focal loss, to address dataset imbalances. It is notable from Table \ref{tab: Table 1} that only the IOU score of the Traffic Sign 
class improved by 0.14 compared to all other models. Our experiments reveal that standard cross-entropy loss is the most effective for training Deformable models on fisheye images. In addition, standard and weighted focal loss yielded promising results in handling dataset imbalances, as presented in \ref{tab: Table 1}. Specifically, the vehicle class achieved a 0.01 improvement in IoU score compared to Vanilla U-Net with standard focal loss. Similarly, with weighted focal loss, the traffic sign class improved by 0.27 and 0.14 in IoU score compared to Vanilla U-Net and Vanilla Deformable U-Net, respectively.

Moreover, the table demonstrates that integrating Deformable Convolutions into U-Net and Residual U-Net models enhances fisheye image segmentation. This underscores the adaptability of Deformable Convolutions to intrinsic fisheye characteristics and geometric transformations. Notably, the best results were achieved by V\_DeU-Net\_ce. Thus, by incorporating Deformable components into encoder and decoder layers, the model can better capture geometric distortions inherent in fisheye imagery and become robust towards camera position variations. Incorporating Deformable layers into the CNN network hinges on the dataset's size. Enhancing the initial and final layers suffices large datasets to capture spatial and geometric distortion effectively.
Conversely, exploring injections across various layers becomes necessary when dealing with smaller datasets. Training on a multi-view dataset has yielded promising results in our scenario, indicating the model's robust performance. Hence, these experiments suggest the potential of deploying a Deformable model for fisheye image segmentation within computational constraints, opening up new avenues for research in this domain.

\begin{table}[htbp]
  \centering
  \scriptsize 
  \captionsetup{font=small}
  \caption{The table presents class-specific accuracy and IoU scores for various configurations of Vanilla and Residual U-Net models, including their deformable versions, trained with different loss functions: cross-entropy (ce), standard focal loss (nwf), and weighted focal loss (wf). The highest IoU score for each class is highlighted in green for improved clarity.}
  
    \begin{adjustbox} {max width=\textwidth}
    \begin{tabular}{p{0.7cm}|p{1.5cm}|cc|cc|cc|cc|cc|cc|cc}
    \toprule
    \textbf{Sr. \#} & \textbf{Categories} & \multicolumn{2}{c|}{\textbf{V\_U-Net\_ce}} & \multicolumn{2}{c|}{\textbf{V\_DeU-Net\_ce}} & \multicolumn{2}{c|}{\textbf{V\_DeU-Net\_nwf}} & \multicolumn{2}{c|}{\textbf{V\_DeU-Net\_wf}} & \multicolumn{2}{c|}{\textbf{R\_U-Net\_ce}} & \multicolumn{2}{c|}{\textbf{R\_DeU-Net\_ce}} & \multicolumn{2}{c}{\textbf{R\_DeU-Net\_wf}} \\
    &  & \textbf{{Acc $\uparrow$}} & \textbf{{IoU $\uparrow$}} & \textbf{{Acc $\uparrow$}} & \textbf{{IoU $\uparrow$}} & \textbf{{Acc $\uparrow$}} & \textbf{{IoU $\uparrow$}} & \textbf{{Acc $\uparrow$}} & \textbf{{IoU $\uparrow$}} & \textbf{{Acc $\uparrow$}} & \textbf{{IoU $\uparrow$}} & \textbf{{Acc $\uparrow$}} & \textbf{{IoU $\uparrow$}} & \textbf{{Acc $\uparrow$}} & \textbf{{IoU $\uparrow$}} \\
    \midrule
    1     & Background & 0.98  & 0.96  & 0.99  & 0.97  & 0.99  & 0.96  & 0.98  & 0.95  & 0.98  & 0.95  & 0.98  & 0.96  & 0.97  & 0.94 \\
    2     & Road  & 0.97  & 0.92  & 0.97  & 0.93  & 0.97  & 0.92  & 0.96  & 0.88  & 0.96  & 0.89  & 0.97  & 0.91  & 0.96  & 0.87 \\
    3     & Lanemark & 0.64  & 0.61 & 0.54  & 0.53  & 0.55  & 0.53  & 0.40  & 0.39  & 0.39  & 0.38  & 0.53  & 0.51  & 0.32  & 0.31 \\
    4     & Curb  & 0.57  & 0.49  & 0.56  & \textcolor{green}{\textbf{0.50}} & 0.55  & 0.48  & 0.53  & 0.48  & 0.38  & 0.36  & 0.46  & 0.42  & 0.47  & 0.43 \\
    5     & Person & 0.28  & 0.23  & 0.37  & \textcolor{green}{\textbf{0.32}} & 0.34  & 0.27  & 0.29  & 0.23  & 0.04  & 0.04  & 0.15  & 0.12  & 0.27  & 0.19 \\
    6     & Rider & 0.44  & 0.41  & 0.54  & \textcolor{green}{\textbf{0.46}} & 0.42  & 0.36  & 0.51  & 0.44  & 0.18  & 0.16  & 0.17  & 0.15  & 0.10  & 0.10 \\
    7     & Vehicles & 0.91  & 0.84  & 0.90  & 0.85  & 0.90  & \textcolor{green}{\textbf0.85} & 0.87  & 0.76  & 0.89  & 0.73  & 0.88  & 0.81  & 0.84  & 0.67 \\
    8     & Bicycle & 0.60  & 0.52  & 0.74  & \textcolor{green}{\textbf{0.53}} & 0.49  & 0.43  & 0.47  & 0.38  & 0.33  & 0.28  & 0.45  & 0.37  & 0.07  & 0.06 \\
    9     & Motorcycle & 0.46  & 0.37  & 0.73  & \textcolor{green}{\textbf{0.47}} & 0.54  & 0.42  & 0.56  & 0.44  & 0.14  & 0.10  & 0.29  & 0.22  & 0.44  & 0.31 \\
    10    & Traffic Sign & 0.09  & 0.09  & 0.11  & 0.12  & 0.10  & 0.10  & 0.39  & \textcolor{green}{\textbf{0.36}} & 0.02  & 0.01  & 0.02  & 0.02  & 0.18  & 0.17 \\
    \midrule
    \multicolumn{2}{c|}{\textbf{Average mIOU}} &       & {\textbf{0.93}} &       & \textcolor{green}{\textbf{0.93}} &       & \textbf{0.91} &       & \textbf{0.88} &       & \textbf{0.89}&          & \textbf{0.89} & & \textbf{0.85} \\
    \multicolumn{2}{c|}{\textbf{Average Accuracy}} & \textbf{0.99} &       & \textbf{0.99} &       & \textbf{0.99} &       & \textbf{0.98} &       & \textbf{0.94} &       & \textbf{0.94} &       & \textbf{0.98} \\
    \bottomrule
    \end{tabular}%
    \end{adjustbox}
  \label{tab: Table 1}%
  \vspace{-0.4cm}
\end{table}%

\section{Conclusion and Future Work}

This study investigated the effectiveness of integrating Deformable Convolutions for semantic segmentation of fisheye images in the automotive domain. We explored four models as a baseline: Vanilla\_U-Net, Residual\_U-Net, Deformable\_U-Net, and Deformable\_Residual\_U-Net. Our proposed approach illustrates the promising potential of Deformable Convolutions in effectively learning fisheye image characteristics. Through our experiments, we observed that integrating Deformable Convolutional blocks allows for more refined and efficient modelling of fisheye images. As a result, future research could explore incorporating these blocks into alternative backbone architectures or multitask networks to enhance the semantic segmentation for synthetic and real-world datasets. Furthermore, avenues for investigation may include tasks such as instance segmentation, detection, and optical flow estimation. Additionally, integrating patchwise mechanisms into the model to capture local and global information about object positions, shapes, or depth maps as constraints could enhance segmentation performance and mitigate challenges associated with class size imbalances within datasets.
We can also integrate this network into a transformer-based architecture for future endeavours to achieve even better semantic segmentation results for the automotive industry.

\section*{Acknowledgments}
This work was supported, in part, by the Science Foundation
Ireland grant 13/RC/2094 P2 and co-funded under the European
Regional Development Fund through the Southern \& Eastern Regional Operational Programme to Lero - the Science Foundation
Ireland Research Centre for Software \textbf{(www.lero.ie)}. 

\bibliographystyle{apalike}

\bibliography{imvip}

\end{document}